\documentclass[12pt,a4paper,twoside]{article}

\usepackage{txfonts}
\usepackage[latin1]{inputenc}
\usepackage[T1]{fontenc}
\usepackage{epsf}
\usepackage{bm}
\usepackage{amsfonts}
\usepackage[small,compact]{titlesec}

\usepackage{taln2006}
\usepackage[frenchb]{babel}
\newcommand{\fup}[1]{\up{#1}}

\renewcommand\sectionmark[1]{}
\renewcommand\subsectionmark[1]{}

\begin{document}

\title{Analyse spectrale des textes~: détection automatique des frontières de langue et de discours}

\author{Pascal Vaillant\fup{1,2}\quad Richard Nock\fup{1}\quad Claudia Henry\fup{1}\\
  (1) GRIMAAG, Université des Antilles-Guyane, 97233 Sch{\oe{}}lcher\\
  (2) GEREC-F, Université des Antilles-Guyane, 97233 Sch{\oe{}}lcher\\
  \{pvaillan,chenry,rnock\}@martinique.univ-ag.fr\\
}

\date{26 f{\'{e}}vrier 2006}

\maketitle

\resume{Nous proposons un cadre théorique qui permet, à partir de
matrices construites sur la base des données statistiques d'un corpus,
d'extraire par des procédés mathématiques simples des informations sur
les mots du vocabulaire de ce corpus, et sur la syntaxe des langues
qui l'ont engendré. À partir des mêmes données initiales, on peut
construire une matrice de similarité syntagmatique (probabilités de
transition d'un mot à un autre), ou une matrice de similarité
paradigmatique (probabilité de partager des contextes
identiques). Pour ce qui concerne la première de ces deux
possibilités, les résultats obtenus sont interprétés dans le cadre
d'une modélisation du processus génératif par chaînes de Markov. Nous
montrons que les résultats d'une analyse spectrale de la matrice de
transition peuvent être interprétés comme des probabilités
d'appartenance de mots à des classes. Cette méthode nous permet
d'obtenir une classification continue des mots du vocabulaire dans des
sous-systèmes génératifs contribuant à la génération de textes
composites. Une application pratique est la segmentation de textes
hétérogènes en segments homogènes d'un point de vue linguistique,
notamment dans le cas de langues proches par le degré de recouvrement
de leurs vocabulaires.}

\abstract{We propose a theoretical framework within which information
on the vocabulary of a given corpus can be inferred on the basis of
statistical information gathered on that corpus. Inferences can be
made on the categories of the words in the vocabulary, and on their
syntactical properties within particular languages. Based on the same
statistical data, it is possible to build matrices of syntagmatic
similarity (bigram transition matrices) or paradigmatic similarity
(probability for any pair of words to share common contexts). When
clustered with respect to their syntagmatic similarity, words tend to
group into sublanguage vocabularies, and when clustered with respect
to their paradigmatic similarity, into syntactic or semantic
classes. Experiments have explored the first of these two
possibilities. Their results are interpreted in the frame of a Markov
chain modelling of the corpus' generative processe(s): we show that
the results of a spectral analysis of the transition matrix can be
interpreted as probability distributions of words within
clusters. This method yields a soft clustering of the vocabulary into
sublanguages which contribute to the generation of heterogeneous
corpora. As an application, we show how multilingual texts can be
visually segmented into linguistically homogeneous segments. Our
method is specifically useful in the case of related languages which
happened to be mixed in corpora.}

\motsClefs{classification spectrale continue, segmentation de textes, identification de langue}
{Soft spectral clustering, text segmentation, language identification}

\newpage
\section{Introduction}

Un corpus nous livre des données brutes sur la syntaxe du système
linguistique utilisé pour le produire. Le mot «~syntaxe~» est employé
ici dans l'acception la plus lâche~: on entend par là l'ensemble des
régularités combinatoires observées, avec une fréquence dépassant un
certain seuil, dans la disposition des unités du corpus. Cette
conception minimaliste, et purement empirique, de la syntaxe, est
d'une part relative à un corpus, et d'autre part limitée à la
description de lois statistiques. L'objet épistémologique de
«~langue~» est, dans ce contexte, hors de notre portée~: dans un
corpus, on peut d'une part avoir plusieurs langues entremêlées, et
l'on n'observe jamais d'autre part {\em toute} la langue, mais
seulement des régularités liées à une situation donnée d'usage. Nous
considérons donc plus prudemment que nous avons affaire, avec un
corpus donné, à un échantillon fourni par un {\em système génératif}~:
celui-ci pouvant être considéré, selon les besoins de la description,
tantôt comme une langue, tantôt comme un type de discours (langage de
spécialité, genre~...), tantôt comme un idiolecte qui peut lui-même
être subordonné à un type de discours.

Depuis Saussure, on envisage les régularités de l'ordre de la syntaxe
sous deux dimensions possibles~: la dimension {\em syntagmatique}, qui
est celle de la disposition d'unités linguistiques dans la chaîne
actualisée par la parole, et la dimension {\em paradigmatique}, qui
est celle de l'ensemble des choix possibles entre unités, en fonction
des contraintes de sélection, à un point donné de la chaîne. Deux
unités linguistiques appartiennent au même {\em syntagme} si elles
sont conjuguées, au sein d'une unité un peu plus ample, dans la même
chaîne parlée~; elles appartiennent au même {\em paradigme} si elles
constituent deux choix possibles pour remplir une position de la
chaîne dans un contexte commun.

En vertu de ces notions classiques en linguistique générale, il est
possible de donner deux définitions de la {\em similarité} syntaxique
de deux unités~: l'une est une similarité syntagmatique~: elle mesure
la probabilité de deux unités de se retrouver associées dans une même
chaîne, à deux positions différentes mais voisines~; l'autre est une
similarité paradigmatique~: elle mesure la probabilité de deux unités
de pouvoir partager le même contexte, c'est-à-dire de pouvoir se
retrouver alternativement à la même position dans des chaînes
contenant par ailleurs les mêmes autres éléments. Selon cette
définition, {\em chien} et {\em aboie} sont très voisins relativement
à une distance syntagmatique, et très éloignés relativement à une
distance paradigmatique. À l'inverse, {\em aboie} et {\em jappe} sont
très voisins relativement à une distance paradigmatique.

En donnant à ces notions des définitions opératoires précises, il est
possible de construire des {\em matrices de dissimilarité} (et
respectivement, des {\em matrices de similarité}) sur l'espace des
mots d'un corpus. L'analyse spectrale de ces matrices (analyse des
vecteurs propres) fournit ensuite des axes de classification des
mots~: une {\em matrice de similarité syntagmatique}, interprétée
comme la matrice de transition d'une chaîne de Markov, nous fournit
les éléments permettant de regrouper ensemble les mots qui sont
souvent voisins les uns des autres sur la même chaîne, et donc d'{\em
identifier les systèmes génératifs}\,. Une {\em matrice de similarité
paradigmatique} nous fournit les éléments permettant de regrouper
ensemble les mots qui partagent souvent le même contexte, et donc
d'{\em identifier les catégories syntaxico-sémantiques}\,. La présente
communication se concentrera sur la première de ces deux analyses, la
seconde étant encore dans une phase d'implémentation plus précoce.

Le système présenté ici se concentre sur la tâche de la classification
des mots-types (vocables) d'un corpus. En regroupant les mots par
proximité syntagmatique, il parvient à des classes de mots qui
reflètent leur appartenance à des systèmes génératifs (langues ou
discours) distincts. Contrairement aux approches comparables fondées
sur la classification de textes \cite{lelu2004}, il n'attribue pas de
classe à chaque texte considéré comme un seul objet individuel~: il
trouve les oppositions entre classes au sein même du corpus qu'il
étudie. Ceci ouvre la possibilité d'une segmentation du texte lui-même
en systèmes génératifs distincts. Par {\em systèmes génératifs}, comme
dit plus haut, nous entendons, selon les cas, les langues, les
discours, les thématiques, les auteurs, ou une combinaison de ces
paramètres\footnote{Pour paraphraser Hjelmslev, notre objectif est de
« donner un système d'étiquettes permettant d'appeler un groupe de
textes ``anglais'', un autre ``danois'', un autre ``prose'', un autre
``poésie'', un autre ``Peter Andersen'' ou ``Lars Petersen'', etc.,
sections qui, on le voit aisément, se croisent de plusieurs
manières »~\cite[p.\/179]{hjelmslev1963a}.}.

L'identification de langue peut être considérée, sous certaines
contraintes, comme un problème résolu~: pour se limiter à la tâche de
l'identification de textes écrits, rédigés dans une seule langue, et
lorsque celle-ci est répertoriée dans une liste de modèles connus, on
sait que la méthode de statistiques sur les $n$-grammes les plus
fréquents est très fiable, et donne des marges d'erreur inférieures à
1\,\% dès que la longueur du texte analysé dépasse quelques dizaines
de caractères. En combinant cette méthode des $n$-grammes avec celle
des mots les plus fréquents, on peut augmenter encore la fiabilité de
l'identification jusqu'à arriver à une marge d'erreur
négligeable~\cite{greffenstette1995}.  Les choses se compliquent
lorsque les textes que l'on cherche à étiqueter sont composites~;
\cite{vo2004}, par exemple, cherche à la fois à segmenter les textes
en zones de langue homogène, et à identifier la langue de chacune de
ces zones. Cependant, dans de telles approches, il est encore
nécessaire de disposer de modèles statistiques des différentes langues
connues, acquis après une phase d'apprentissage supervisé. Or nous
partons de données brutes, sans aucun modèles préalable. Nous
cherchons, dans le cadre de ce travail, à identifier, au sein d'un
texte, des langues différentes mais (a)~fréquemment entremêlées dans
des textes, et (b)~proches, au sens où elles partagent une grande
quantité d'unités lexicales. Le but est de démêler des langues qui
sont proches à la fois génétiquement et
« sociolinguistiquement ». C'est notamment le cas dans la littérature
antillaise, pour le français et le créole.

Des approches statistiques ont été appliquées à la segmentation
automatique de texte pour identifier des segments
homogènes~\cite{choi2000}, sur d'autres plans que celui de la langue,
comme par exemple sur le plan du thème ou sur celui de
l'auteur. \cite{utiyama-isahara2001} utilisent notamment une méthode
qui définit les frontières des segments de façon à maximiser leur
homogénéité interne. \cite{elbeze-torres-bechet2005} introduisent le
modèle de la chaîne de Markov pour modéliser les transitions entre
segments de texte provenant du même auteur.

Des travaux plus proches des nôtres sont celui de
\cite{belkin-goldsmith2002}, qui utilisent une forme de classification
spectrale pour déterminer des catégories de mots~; et celui de
\cite{pessiot-et-al2004}, qui commencent par classer les mots du
corpus et se servent de cette classification pour identifier des
segments. Cependant, dans ce travail, la méthode de classification
fait le choix délibéré de n'assigner chaque mot qu'à un cluster et un
seul. Pour notre application, nous avons besoin d'une méthode qui
puisse à la fois identifier les principaux regroupements spécifiques,
tout en gardant également l'information concernant les mots qui
appartiennent à la fois à plusieurs clusters.

Notre méthode de classification spectrale nous offre la possibilité
d'une classification {\em continue}, c'est-à-dire qu'elle n'attribue
pas de manière discrète et univoque un mot à une classe, mais
quantifie la probabilité d'appartenance d'un mot à chacune des
classes. Enfin, une originalité de ce travail réside dans une
interprétation probabiliste des coordonnées des axes propres des
matrices de similitude.

\section{Éléments du modèle}

Nous partons du postulat que nous disposons d'un corpus écrit
(i.e. d'une production langagière déjà discrétisée en éléments
d'expression élémentaires~--- des caractères)~; et, en outre, que ce
corpus est transcrit dans un système d'écriture qui possède déjà une
convention de délimitation des unités lexicales.  C'est le cas dans
les langues dont nous avons étudié des textes, qui sont des langues
vivantes transcrites en caractères latins~: les langues de cette
catégorie ont à notre connaissance généralisé la convention consistant
à délimiter les mots par des caractères spéciaux (signes de
ponctuation, ou au moins caractère d'espacement). La méthode pourrait
être généralisée aux langues n'appliquant pas cette convention~---
telles le chinois~--- au prix d'un prétraitement de délimitation des
unités lexicales.

Dans ce cadre, nous disposons des catégories pré-théoriques
suivantes~: le {\em texte} est une séquence de caractères~; le {\em
mot-occurrence} est défini «~mécaniquement~», au sein d'un texte,
comme une séquence de caractères alphabétiques consécutifs, délimitée
à gauche et à droite par des caractères non-alphabétiques.  On peut
choisir d'inclure, dans une définition étendue de «~mot~» (les {\em
tokens}), les séquences de chiffres et les signes de ponctuation
également représentés en un ou plusieurs exemplaires dans le texte~:
ce choix est un paramètre d'usage qu'il peut être utile de fixer au
dernier moment, en fonction des informations attendues des
résultats\footnote{Pour la délimitation des classes syntaxiques au
sein d'une langue, il peut être parfaitement pertinent d'inclure les
virgules et les points, qui permettent par exemple de mettre
facilement le doigt sur les classes systématiquement présentes en
début de phrase~; cette donnée aurait au contraire plutôt tendance à
brouiller les choses lors de la délimitation de différentes langues,
utilisant toutes deux les mêmes signes de ponctuation.}. Pour
l'instant, nous utilisons le terme {\em mot-occurrence} au sens le
plus général, sans nous préoccuper de savoir si nous y incluons les
«~mots~» non-alphabétiques ou non. Les notions suivantes découlent des
premières~: un {\em mot-type} (ou {\em vocable}) est un agencement de
caractères alphabétiques qui se retrouve à une ou plusieurs reprises
en tant que mot-occurrence dans les textes. Un {\em corpus} $\cal{C}$
est un ensemble de textes. Le {\em vocabulaire} d'un corpus,
$\cal{V}(\cal{C})$ est l'ensemble des mots-types représentés par au
moins une occurrence dans le corpus.

Pour simplifier, nous assimilons dans la suite le corpus à la
concaténation de tous les textes qui le composent, bien que les deux
notions ne soient pas identiques (lors de la collecte d'informations
statistiques sur un corpus, on ne compte en effet pas le dernier mot
du texte $T_{k}$ comme « voisin de gauche » du premier mot du texte
$T_{k+1}$).  Nous parlons donc dorénavant indifféremment du « texte »
comme s'il constituait à lui seul tout le corpus, ce qui simplifie
l'exposé des principes de notre travail, sans en modifier
fondamentalement le sens.

Soit un texte $T$ constitué d'une séquence de $no$ occurrences
$w_{1}w_{2}...w_{no}$, choisies parmi un vocabulaire
$\mathcal{V}(T)=\{m_{1},m_{2},...m_{nt}\}$ ($nt \leq{} no$). D'un
point de vue algébrique, un texte est ni plus ni moins qu'une
application surjective $f_{T}$ de $[1,no] \subset \mathbb{N}$ dans
$[1,nt] \subset \mathbb{N}$~: on note $f_{T}(j) = i$ lorsque $w_{j}$
est une occurrence du mot $m_{i}$.

\subsection{Matrices}
\label{matrices}

Nous définissons la {\em matrice de contexte} à la distance $k$ ($k
\in \mathbb{Z}$), notée $C_{k}$, comme une matrice de nombres entiers,
de dimension $nt \times nt$, dans laquelle chaque coefficient
$c_{k}(i,j)$ correspond au nombre de fois où une occurrence du mot $j$
apparaît à la position $p+k$ lorsqu'une occurrence du mot $i$ apparaît
à la position $p$ (pour tout $p \in [1,no]$). Par exemple, pour $k =
+1$, cette matrice correspond à la matrice de {\em contexte immédiat à
droite}~: le coefficient de $C_{+1}$ à la ligne $i$ et à la colonne
$j$, $c_{+1}(i,j)$, contient le nombre de fois où le mot $j$ s'est
retrouvé juste après le mot $i$ dans le texte. De même, $C_{-1}$
correspond à la matrice de {\em contexte immédiat à gauche}, $C_{+2}$
à la matrice de contexte à droite à deux mots de distance,
etc. Conformément à cette définition, $C_{0}$ est une matrice qui
contient des zéros partout sauf sur la diagonale, et dont les
coefficients diagonaux contiennent le nombre d'occurrences de chaque
mot. Dans la suite de cet exposé, pour simplifier la notation, nous la
noterons $D$~: $D = C_{0}$, et nous noterons en abrégé $d_{i}$ ses
coefficients diagonaux~: $d_{i} = c_{0}(i,i)$ correspond au nombre
d'occurrences du mot $i$ dans le texte. Il convient par ailleurs de
noter que quel que soit $k$, la somme de tous les coefficients de la
$i$-ième ligne de la matrice $C_{k}$, ainsi que la somme de tous les
coefficients de la $i$-ième colonne de cette même matrice, sont
toujours égales à $d_{i}$~: en effet, les coefficients stockés sur la
$i$-ième ligne correspondent à l'ensemble des fois où le mot $i$ a eu
tel ou tel voisin à la distance $k$ (donc il y en a au total autant
que d'occurrences de $m_{i}$ dans le texte)~; de la même manière, les
coefficients de la $i$-ième colonne correspondent à l'ensemble des
fois où un autre mot a eu le mot $i$ comme voisin à la distance $k$.

Dans le travail présenté ici, nous cherchons à extraire
automatiquement des informations des corpus sans faire au départ
aucune hypothèse sur la langue~--- ou le « système génératif »~---
dans lequel ceux-ci ont été composés. Ceci implique notamment de nous
affranchir de toute hypothèse sur la directionalité de la lecture.
Dans cet objectif, nous définissons un mode de lecture « circulaire »
de chaque texte~: nous supposons que le texte nous est donné comme un
espace linéaire statique qu'un « lecteur » peut indifféremment (et
aléatoirement) parcourir de gauche à droite ou de droite à gauche.
Afin de prendre en compte cette symétrisation, nous définissons une
{\em matrice de voisinage} à la distance $k$ ($k \in \mathbb{Z}$),
notée $V_{k}$, comme une matrice de nombres entiers, de dimension $nt
\times nt$, dans laquelle chaque coefficient $v_{k}(i,j)$ correspond
au nombre de fois où une occurrence du mot $i$ et une occurrence du
mot $j$ se retrouvent à une distance de $k$ mots l'une de l'autre dans
la chaîne~: $v_{k}(i,j)=c_{k}(i,j)+c_{k}(j,i)$. En pratique, pour
conserver une homogénéité de norme avec $C$, il est préférable de
diviser ces coefficients par un facteur $2$ (dans $V$ chaque
occurrence est en effet comptée deux fois)~; pour chaque $k$, nous
définissons donc une nouvelle matrice $W_{k}$ comme la matrice de
dimension $nt \times nt$ et de coefficients~: $w_{k}(i,j) =
\frac{1}{2}v_{k}(i,j) = \frac{1}{2}\Big(
c_{k}(i,j)+c_{k}(j,i)\Big)$. Ainsi, les sommes par ligne et par
colonne de $W$ sont elles aussi égales aux coefficients $d_{i}$.  La
matrice $W$ est l'équivalent de la matrice de contexte, mais avec un
sens de lecture rendu aléatoire~: c'est une matrice de contexte
équidirectionnelle.

L'étape suivante est de ramener ces informations, extraites d'un
corpus donné, à un modèle abstrait indépendant de la taille du texte.
Pour cela, il faut normaliser les informations sur le nombre de cas de
voisinages observés autour du mot $i$, en les divisant par le nombre
d'occurrences de ce mot. On définit donc une {\em matrice de
probabilité de transition} à la distance $k$, notée $P_{k}$, comme
étant la matrice de nombres réels, de dimension $nt \times nt$, et
dont les coefficients $p_{k}(i,j)$ correspondent à la probabilité
qu'une occurrence du mot $m_{j}$ soit observée à une position $p \pm
k$ sachant qu'une occurrence du mot $m_{i}$ a été observée à la
position $p$~: $p_{k}(i,j) = \frac{1}{d_{i}} w_{k}(i,j)$. En termes de
multiplication de matrices, la matrice $P$ se définit donc globalement
par~: $P_{k} = D^{-1} W_{k}$. La matrice $P_{k}$ ainsi définie a un
certain nombre de propriétés intéressantes qui sont détaillées en
\S{}\,\ref{similarites} et \S{}\,\ref{markov}.

Par ailleurs, avec les informations de base tirées des matrices de
contexte $C_{k}$, on peut construire des matrices de dissimilarité et
de similarité paradigmatiques entre les mots de
$\mathcal{V}(\mathcal{C})$.

Tout d'abord, on définit pour chaque position contextuelle $k$ une
grandeur appelée {\em dissimilarité des contextes du mot $i$ et du mot
$j$ à la position $k$}, qui correspond à la proportion de mots
spécifiques au contexte à la position $k$ de $i$ ou à celui de $j$,
rapportée au nombre total de mots pouvant apparaître dans les
contextes à la position $k$ de l'un aussi bien que de l'autre~:

\begin{eqnarray*}
\delta{}^{c}_{k}(i,j) =
\frac{\sum_{x=1}^{nt} |c_{k}(i,x)-c_{k}(j,x)|}
     {\sum_{x=1}^{nt} (c_{k}(i,x)+c_{k}(j,x))}
=
\frac{\sum_{x=1}^{nt} |c_{k}(i,x)-c_{k}(j,x)|}
     {d_{i}+d_{j}}
\end{eqnarray*}

Pour prendre l'exemple de $k=+1$~: $\delta{}^{c}_{+1}(i,j)$ correspond
au cardinal de la différence symétrique des contextes immédiats à
droite de $m_{i}$ et de $m_{j}$, divisé par le cardinal de l'union de
ces deux contextes (ou encore~: au nombre de mots-occurrences
apparaissant seulement derrière $m_{i}$ ou seulement derrière $m_{j}$,
divisé par le nombre total de mots-occurrences apparaissant derrière
$m_{i}$ ou $m_{j}$~--- c'est-à-dire divisé par la somme du nombre
d'occurrences de $m_{i}$ et du nombre d'occurrences de $m_{j}$).  Dans
le cas particulier de $k=0$~: $\delta{}^{c}_{0}$ se ramène à~: \(
\delta{}^{c}_{0}(i,j) = 0 \textrm{ si $i=j$} \)~; \(
\delta{}^{c}_{0}(i,j) = 1 \textrm{ si $i \neq j$} \).

On peut ensuite définir une {\em dissimilarité contextuelle} (ou {\em
dissimilarité paradigmatique}) prenant en compte les contextes d'un
intervalle $I=[k_{1},k_{2}]$, comme une somme pondérée des
dissimilarités contextuelles $\delta{}^{c}_{k}$ pour tous les $k \in
I$. Par exemple si les poids valent 1, $\delta{}^{c}(i,j) = \sum_{k
\in I}\delta{}^{c}_{k}(i,j)$.

Une matrice de {\em similarité contextuelle} S peut alors être
construite en transformant la dissimilarité $\delta{}^{c}$ en
similarité, par la formule simple $s(i,j)=1-\delta{}^{c}(i,j)$
($\delta{}^{c}(i,j)$ étant comprise dans $[0,1]$, $s(i,j)$ l'est
aussi).

\subsection{Dissimilarités et similarités}
\label{similarites}

Il est possible de montrer que $\delta{}^{c}$, définie comme ci-dessus
(\S{}\,\ref{matrices}), est une véritable distance sur
$\mathcal{V}(\mathcal{C})$~: pour commencer, elle prend ses valeurs
sur $\mathbb{R}^{+}$, et elle est symétrique.  L'inégalité
triangulaire est démontrable, mais nous n'en inclurons pas la preuve
ici faute de place. Enfin, si l'intervalle $I$ considéré ci-dessus
contient $k=0$, alors $\delta{}^{c}$ vérifie également l'axiome de
séparation ($\forall i,j \in \mathcal{V}(\mathcal{C}),
\delta{}^{c}(i,j)=0 \Leftrightarrow i=j$), puisque même si deux
vocables distincts $m_{i}$ et $m_{j}$ ont exactement les mêmes
contextes sur les autres positions $k \in I$, la définition de
$\delta{}^{c}_{0}(i,j)$ suffit à les discriminer. Cette distance sur
l'espace des mots ouvre la voie à différentes possibilités
d'algorithmes de minimisation. En outre, il est intéressant de noter
que cette distance possède une interprétation en termes bayésiens~:
$\delta{}^{c}(i,j)$ correspond en effet à la probabilité de se trouver
dans un contexte qui soit spécifique à l'un des deux mots $m_{i}$ ou
$m_{j}$, sachant que l'on est dans une position contextuelle qui
permet au moins l'un des deux, et peut-être les deux~: bref,
$\delta{}^{c}(i,j)$ représente la probabilité que le contexte courant
{\em discrimine} $m_{i}$ et $m_{j}$ au sein de leur paradigme commun.
Cette interprétation a des implications en termes de définition de
catégories syntaxiques et sémantiques que nous n'avons pas fini
d'explorer.

Revenons maintenant à la matrice de probabilité de transition $P$,
définie plus haut dans \S{}\,\ref{matrices}. Dans la suite, pour
alléger les notations, nous considérons le cas particulier où $k=1$,
et sauf indication contraire, nous notons simplement $P$ pour $P_{1}$
et $W$ pour $W_{1}$. Ce qui en est dit est généralisable sans
difficulté aux autres valeurs possibles de $k$.

Tout d'abord, on peut observer que les coefficients de $P$, les
$p(i,j)$, possèdent certaines propriétés d'une {\em fonction de
similarité} sur l'ensemble des mots~: c'est une fonction de
$\cal{V}(\cal{C}) \times \cal{V}(\cal{C})$ dans $]0,1] \subset
\mathbb{R}^{+}$. On peut définir à partir de cette fonction une
fonction de {\em dissimilarité}, grâce à une transformation du type
$\delta{}(i,j) = 1-p(i,j)$, ou $\delta{}(i,j) = ( 1-p(i,j) ) /
p(i,j)$, définie de $\cal{V}(\cal{C}) \times \cal{V}(\cal{C})$ dans
$\mathbb{R}^{+}$, qui vérifie l'axiome de séparation ($\delta{}(i,j) =
0 \Leftrightarrow i=j$) et l'axiome de symétrie ($\forall
i,j,$~$\delta{}(i,j) = \delta{}(j,i)$). Cependant cette fonction ne
vérifie pas l'inégalité triangulaire\footnote{On ne peut pas affirmer,
connaissant un mot $m_{x}$ et un mot $m_{y}$, qu'il est impossible de
trouver un mot $m_{z}$ tel que $p(x,z)+p(z,y)>p(x,y)$, en d'autres
termes tel que la probabilité de transition de $m_{x}$ à $m_{y}$ {\em
en passant par $m_{z}$} soit plus grande que la probabilité de passer
directement de $m_{x}$ à $m_{y}$. Ce cas de figure est au contraire
tout à fait courant~: dans le composé {\em match de barrage} (exemple
emprunté à Véronis),
$p(\mathit{match},\mathit{de})+p(\mathit{de},\mathit{barrage})>p(\mathit{match},\mathit{barrage})$.}
qui en ferait une {\em distance} à proprement parler, et ne peut donc
pas directement servir de critère de base à un algorithme de
minimisation. Véronis~\shortcite{veronis2003}, qui utilise une
fonction de similarité proche de la nôtre pour cartographier la
proximité syntagmatique de mots sur de grands corpus, attaque le
problème par des heuristiques de détection de composantes fortement
connexes dans des grands graphes de mots co-occurrents. Notre approche
est différente et fait appel à des méthodes de classification
spectrale continue, en partant d'une autre propriété intéressante de
$P$, dont il va être question en \S{}\,\ref{markov}.

\section{Interprétation en termes de chaînes de Markov}
\label{markov}

La matrice $P$ constitue, d'après un lemme classique (cf. par exemple
\cite{petruszewycz1981}), l'{\em estimation par maximum de
vraisemblance}, sur l'ensemble des bigrammes observés, de la matrice
de transition de la chaîne de Markov engendrant le texte $T$.

\begin{figure}
\begin{center}
\epsfxsize=37mm
\mbox{\epsffile{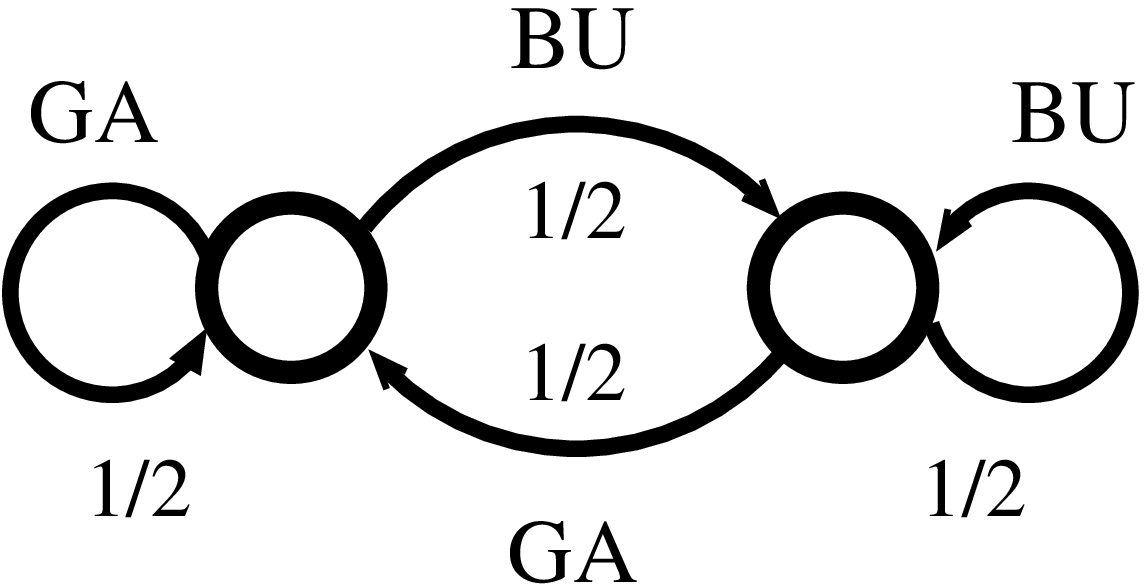}}
\makebox[10mm]{}
\epsfxsize=37mm
\mbox{\epsffile{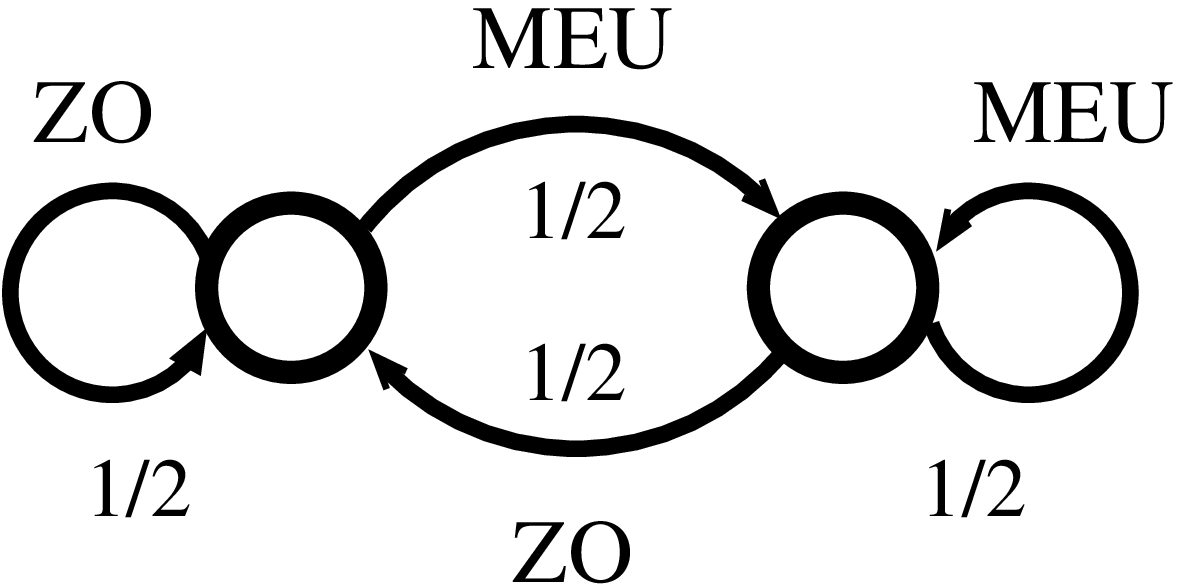}}
\makebox[10mm]{}
\epsfxsize=61mm
\epsfysize=40mm
\mbox{\epsffile{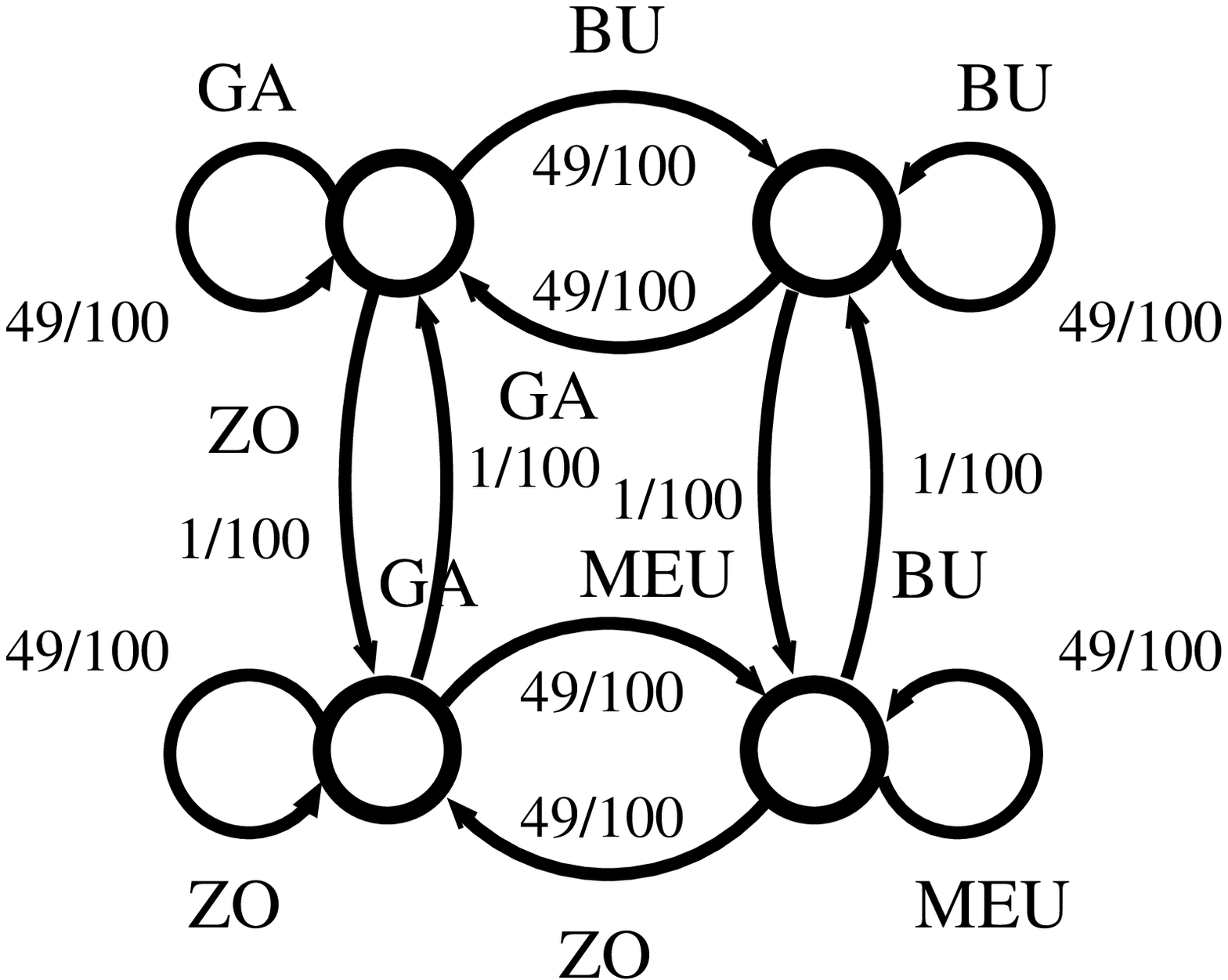}}
\leavevmode
\caption[gabu-zomeu]{\small\sl À gauche, deux exemples de processus
génératifs élémentaires~: (a)~l'un produit des chaînes comme~: «~GA BU
BU BU BU GA BU GA BU BU BU BU BU BU BU BU GA GA GA BU~»~; (b)~l'autre
des chaînes comme~: «~ZO ZO ZO MEU ZO MEU ZO MEU MEU ZO MEU MEU MEU
MEU MEU ZO MEU ZO ZO ZO~». À droite~: (c)~Processus génératif mixte
produisant des chaînes comme~: «~BU GA GA BU BU GA BU BU GA ZO MEU MEU
ZO MEU MEU ZO MEU MEU ZO ZO~»~; la probabilité que la «~parole~» passe
de l' «~auteur~» disant «~GA-BU~» à l' «~auteur~» disant «~ZO-MEU~»
est de 2\,\%.}
\vspace{-4mm}
\label{gabu-zomeu}
\end{center}
\end{figure}

Revenons un instant sur la notion que nous cherchons à appréhender en
parlant de {\em système génératif}. Comme nous l'avons dit plus haut,
il peut s'agir d'une langue, d'un genre, d'un idiolecte~... D'un point
de vue empirique, nous n'avons pas d'autre moyen de caractériser cette
variable que les $n$-grammes que nous observons dans le corpus. Les
chaînes de Markov peuvent donc se révéler un outil de modélisation
tout à fait adapté pour la représenter. Pour simplifier, nous
considérons pour l'instant uniquement les probabilités des
bigrammes\footnote{Dans ce contexte, bigramme signifie~: «~groupe de
deux {\em mots} consécutifs~».}. Dans ce cadre, il est facile de se
représenter ce que signifie un système génératif~: il suffit de
s'imaginer un «~auteur~» comme un agent simpliste effectuant un
parcours aléatoire le long des transitions d'une chaîne de Markov. Le
résultat du parcours est la chaîne engendrée. Nous donnons
fig.~\ref{gabu-zomeu} (a) et (b) deux exemples de ce que peuvent
produire des «~auteurs~» simplistes parcourant de telles chaînes de
Markov.

Un texte composite peut alors être conçu comme le résultat du parcours
aléatoire d'une chaîne de Markov complexe, résultant d'une combinaison
de plusieurs chaînes de Markov plus compactes et plus homogènes, comme
illustré dans la fig.~\ref{gabu-zomeu} (c). Le but de notre tâche de
classification peut alors être conçu comme un travail de {\em
séparation} des deux processus génératifs, qui fonctionne en
attribuant à chaque état une probabilité d'appartenance à l'une ou
l'autre des sous-chaînes.

Notre méthode de classification se fonde ensuite sur le principe de la
classification {\em spectrale}, c'est-à-dire fondée sur la répartition
des mots le long des principaux axes propres de la matrice
$P$. Remarquons tout d'abord que $P$, sans être symétrique, est
définie par le produit de deux matrices réelles symétriques ($D^{-1}$
et $W$, cf.  \S{}\,\ref{matrices}), et qu'elle admet donc des valeurs
propres réelles. D'autre part, par construction, $P$ est {\em
stochastique par lignes}, c'est-à-dire que la somme de tous les
coefficients d'une même ligne est toujours égale à 1 (c'est la somme
des probabilités de transition d'un mot $m_{i}$ à un mot suivant).
Ceci lui confère une autre propriété, qui est d'être de norme 1, et
d'admettre également 1 comme plus grande valeur propre. Les autres
valeurs propres appartiennent toutes à $]0,1[$ et peuvent être (sans
perte de généralité) considérées comme classées par ordre
décroissant~: $\lambda{}_{2}$, $\lambda{}_{3}$, etc. (il y en a $nt$
en tout).

Il convient de noter que le problème de la recherche d'une grande
valeur propre de $P$ est équivalent à celui de la recherche d'une
valeur propre proche de zéro pour le problème de valeurs propres
généralisé $(D-W)\bm{y}=\mu{}D\bm{y}$ (à une multiplication par
$D^{-1}$ près, et en notant $\mu{}=1-\lambda{}$). Ce problème résulte
de la réécriture de la contrainte de minimisation d'un {\em critère de
coût normalisé}, utilisé en classification pour modéliser la recherche
d'un minimum de conductance entre classes. La conductance entre une
classe $k$ et son complémentaire y est évaluée comme la somme, pour
toutes les paires de mots $(i,j)$, d'une distance représentant la
probabilité d'appartenance commune à la classe $k$, pondérée par le
poids de la transition possible entre $i$ et $j$~: $\kappa_k(Z) =
\sum_{i,j=1}^{v} {w_{ij} (z_{ik} - z_{jk})^2}$ ($z_{ij}$ dénote la
fonction caractéristique de l'appartenance du mot $i$ à la classe
$k$). Le critère de coût normalisé a notamment été introduit par
\cite{shi-malik2000} pour des applications de segmentation d'image, et
également utilisé par \cite{belkin-goldsmith2002} pour le clustering
de mots français et anglais. La classification spectrale discrète
procède ensuite à la recherche d'une fonction de seuil~; dans notre
application, nous nous intéressons à une classification continue, donc
nous allons utiliser directement les vecteurs propres de $P$.

Les axes propres d'une matrice peuvent être caractérisés par des
vecteurs de norme quelconque, pourvu qu'ils soient sur le bon axe. Les
fonctions de calcul d'algèbre linéaire que nous utilisons (paquetage
{\sc lapack}) fournissent des vecteurs propres $\bm{y}'_{k}$
normalisés de telle sorte que $\|\bm{y}'_{k}\|_{2}=1$. Notons
$y'_{k}(i)$ les $nt$ coordonnées des vecteurs $\bm{y}'_{k}$. Soient
maintenant $\bm{y}_{k}$ ($k \in [1,nt]$) les vecteurs propres de $P$
définis par multiplication des $\bm{y}'_{k}$ par une constante
scalaire définie pour chacun d'entre eux par
$1/\sqrt{\sum_{i=1}^{nt}d_{i}y'_{k}(i)^{2}}$. Si nous notons
$y_{k}(i)$ les $nt$ coordonnées des vecteurs $\bm{y}_{k}$, ces
vecteurs vérifient par construction~:
$\sum_{i=1}^{nt}d_{i}y_{k}(i)^{2} = 1$ (ou encore~:
$\bm{y}^{\top{}}D\bm{y} = \bm{1}$). Ceci signifie que
$d_{i}y_{k}(i)^{2}$ a les propriétés d'une {\em distribution de
probabilité} sur $\mathcal{V}(\mathcal{C})$.

En tant que vecteurs propres, ces $\bm{y}_k$ vont être caractérisés
par des coordonnées positives ou négatives qui correspondent à une
corrélation de chaque mot avec une composante principale donnée~;
cependant nous voulons insister sur l'autre interprétation possible
des vecteurs propres $\bm{y}_{k}$, qui est spécifique à l'approche
présentée. Nous pouvons considérer chaque axe propre comme une classe
(appelons-la $\mathcal{L}_{k}$), chaque classe comportant {\em tous}
les mots, mais avec une «~probabilité d'appartenance~» qui est donnée,
pour chaque mot $m_{i}$, par $d_{i}y_{k}(i)^{2}$. Sous ce point de
vue, la grandeur $d_{i}y_{k}(i)^{2}$ s'interprète donc comme la {\em
probabilité d'observer le mot $m_{i}$ sachant qu'on est dans la classe
$\mathcal{L}_{k}$}~: $\mathbf{Pr}(m_{i}|\mathcal{L}_{k})$. Il est
intéressant d'observer au passage que le premier vecteur propre
$\bm{y}_{1}$, généralement rejeté comme trivial, correspond en fait
dans cette interprétation à la classe globale du corpus observé,
$\mathcal{L}_{1}$, contenant tous les mots avec une probabilité qui
est précisément leur probabilité d'apparition dans le corpus
($y_{1}(i)=d_{i}/\sum_{i=1}^{nt}d_{i}$).  Ensuite, pour $k>1$, nous
voyons se former des classes qu'il est possible d'interpréter comme
représentant l'appartenance de chaque mot à tel ou tel système
génératif.

\section{Applications et Conclusion}
\label{applications}

\begin{figure}
\begin{center}
\epsfxsize=125mm
\mbox{\epsffile{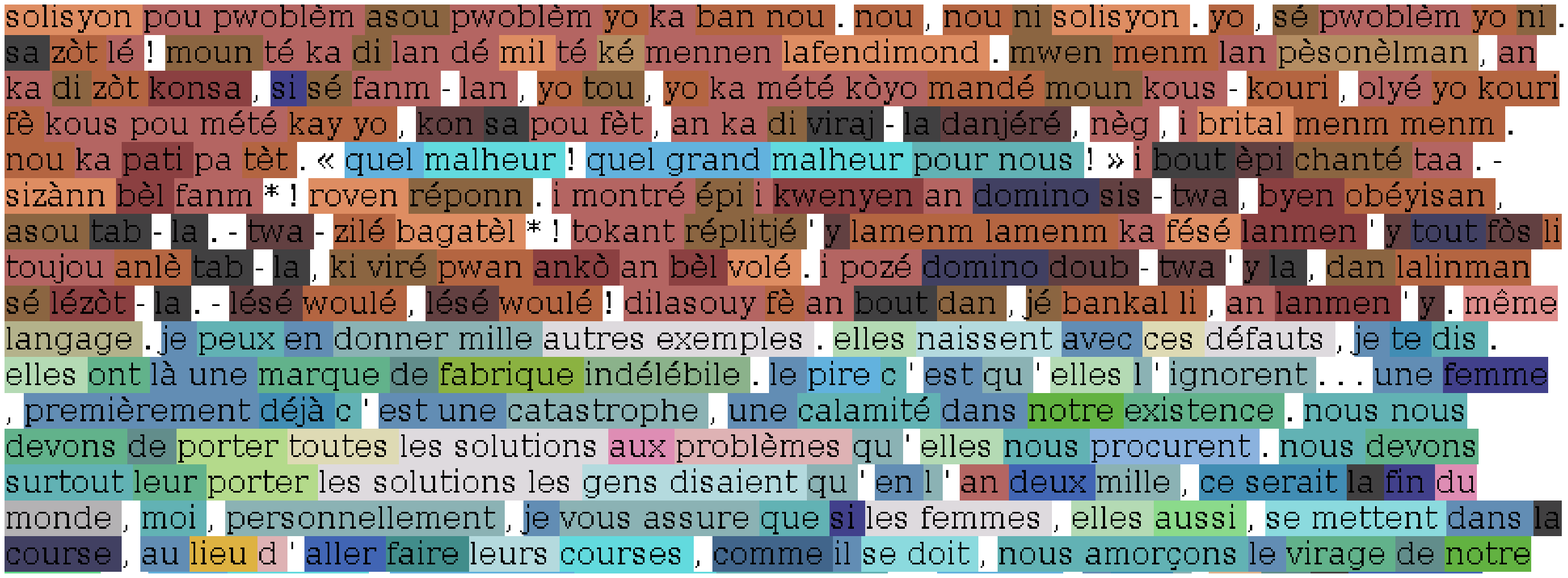}}
\leavevmode
\caption[lavwa_egal_C]{\small\sl Une visualisation par codes de
couleur de l'appartenance de chaque mot aux classes correspondant aux
axes propres de rang 2, 3 et 4, sur l'exemple d'un texte bilingue
français et créole martiniquais (roman {\em Lavwa Egal~--- La voix
égale}, de Thérèse Léotin, 2003). Sur ce schéma, les grandeurs
représentées sont proportionnelles aux {\em carrés} des coordonnées de
chaque mot sur les vecteurs propres.}
\label{lavwa_egal_C}
\vspace{-6mm}
\end{center}
\end{figure}

Nous avons utilisé les méthodes décrites plus haut pour classifier
automatiquement les mots de corpus composites comportant des
échantillons de plusieurs langues entremêlées. Nous nous sommes en
particulier intéressés au cas de textes bilingues en français et en
créole. Afin de mieux visualiser les résultats, nous avons adopté une
méthode consistant à faire ré-afficher tous les mots dans l'ordre du
texte, en les faisant apparaître sur un fond coloré dont le codage
vectoriel RVB correspond à l'affichage des coordonnées du vocable sur
trois axes propres significatifs (les meilleurs résultats étant en
général donnés par le choix des axes propres correspondant aux valeurs
propres $\lambda{}_{2}$, $\lambda{}_{3}$ et $\lambda{}_{4}$). Un
exemple de résultat de ce type de visualisation est donné
fig.~\ref{lavwa_egal_C}. Notons que ce procédé ne permet pas de
visualiser des {\em langues} à proprement parler, puisque le système a
créé des classes de manière totalement non supervisée, et que ce sont
ces classes qui servent de base à l'affichage~; pour attribuer une
{\em langue}, choisie dans un ensemble prédéfini, à chaque mot du
corpus, il faut effectuer une deuxième étape d'apprentissage,
supervisée cette fois, qui nécessite l'étiquetage préalable de
quelques mots représentatifs de chacune des langues manifestées.

\begin{figure}
\begin{center}
\epsfysize=72mm
\mbox{\epsffile{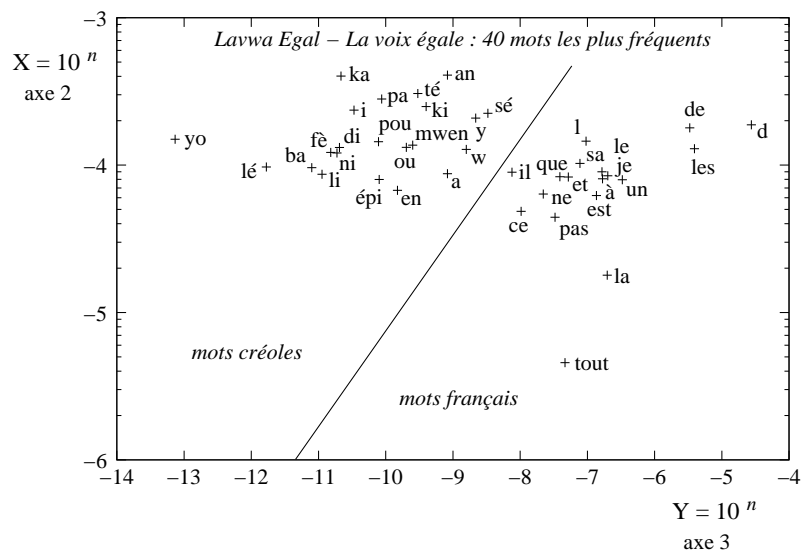}}
\leavevmode
\end{center}
\vspace{-6mm}
\caption[lavwa_egal_40pf]{\small\sl Les 40 mots les plus fréquents de
{\em Lavwa egal~--- La voix égale}\,, développés sur les deux premiers
axes principaux de valeur propre~<~1\,. N.B. ``en'', ``y'', ``an'',
``tout'' et ``la'' sont tout à la fois des mots créoles et des mots
français (parmi eux, ``y'', ``an'' et ``tout'' apparaissent plus
fréquemment dans des contextes de phrases créoles que dans des
contextes de phrases françaises).}
\label{lavwa_egal_40pf}
\end{figure}

Les positions respectives des mots les plus fréquents du corpus sur
les deux premiers axes propres peuvent également être représentées sur
un diagramme bidimensionnel comme celui de la
fig.~\ref{lavwa_egal_40pf}.

Les méthodes mises en {\oe{}}uvre dans ce travail ont prouvé leur
efficacité sur des problèmes génériques, en se limitant volontairement
à des données de base extrêmement simples (matrices de transition sur
des bigrammes). Ils arrivent à identifier des sous-systèmes génératifs
à partir de corpus bruts, sans aucune information enrichie. La
principale voie d'amélioration que nous comptons explorer est celle de
l'enrichissement mutuel des deux approches présentées (catégorisation
syntagmatique et paradigmatique) dans le cadre d'un processus
itératif, afin d'aller vers l'annotation automatique ou
semi-automatique des corpus~--- cette annotation servant elle-même de
base à une meilleure segmentation. L'objectif est d'arriver à une
plate-forme d'acquisition automatique d'information sur les langues ou
sur les genres.

\bibliographystyle{taln2002}
\bibliography{biblio-vnh}


\end{document}